%% file: main.tex
\documentclass[11pt,a4paper]{article}
\usepackage[hyperref]{acl2020}
\usepackage{times}
\usepackage{latexsym}

\usepackage{microtype}

\aclfinalcopy 
\def\aclpaperid{1945} 

\usepackage{verbatim}
\usepackage{amsmath}
\usepackage{amsfonts}
\usepackage{url,placeins}
\usepackage{todonotes}
\usepackage{xcolor}
\definecolor{maskhighlight}{RGB}{140,220,255}
\newcommand{\mask}[1]{\colorbox{maskhighlight!50}{\vphantom{Iq}#1}}
\newcommand{\answer}[1]{\colorbox{maskhighlight!50}{\vphantom{Iq}#1}}
\newcommand{\bertbase}{BERT$_{\text{BASE}}$}
\newcommand{\bertlarge}{BERT$_{\text{LARGE}}$}
\newcommand{\best}[1]{\textbf{#1}}

\definecolor{darkgreen}{rgb}{0.0, 0.2, 0.13}

\newcommand{\colorann}[3]{}

\newcommand{\avi}[1]{\colorann{red}{Avi}{#1}}
\newcommand{\mrglass}[1]{\colorann{violet}{MGlass}{#1}}

\usepackage{booktabs}       
\usepackage{makecell}
\title{Span Selection Pre-training for Question Answering}
\author{
Michael Glass,\textsuperscript{\rm 1} 
Alfio Gliozzo,\textsuperscript{\rm 1} 
Rishav Chakravarti,\textsuperscript{\rm 1}
Anthony Ferritto,\textsuperscript{\rm 1} \\
\textbf{Lin Pan,\textsuperscript{\rm 1} 
G P Shrivatsa Bhargav,\textsuperscript{\rm 2} 
Dinesh Garg,\textsuperscript{\rm 1} 
Avirup Sil\textsuperscript{\rm 1}}\\ 
\textsuperscript{\rm 1} IBM Research AI\\
\textsuperscript{\rm 2} Dept. of CSA, IISC, Bangalore\\ 
\texttt{mrglass@us.ibm.com, gliozzo@us.ibm.com, }\\
\texttt{rchakravarti@us.ibm.com, aferritto@ibm.com, panl@us.ibm.com, }\\
\texttt{bhargavs@iisc.ac.in, garg.dinesh@in.ibm.com, avi@us.ibm.com}\\
}
 
\begin{document}

\maketitle

\begin{abstract}
BERT (Bidirectional  Encoder Representations  from  Transformers) and related pre-trained Transformers have provided large gains across many  language  understanding  tasks, achieving a new state-of-the-art (SOTA).
BERT is pre-trained on two auxiliary tasks: Masked Language Model and Next Sentence Prediction. 
In this paper we introduce a new pre-training task inspired by reading comprehension 
to better align the pre-training from memorization to understanding.
{\em Span Selection Pre-Training} (SSPT) poses cloze-like training instances, but rather than draw the answer from the model's parameters, it is selected from a relevant passage.
We find significant and consistent improvements over both \bertbase~and \bertlarge~on multiple Machine Reading Comprehension (MRC) datasets. 
Specifically, our proposed model has strong empirical evidence as it obtains SOTA results on Natural Questions, a new benchmark MRC dataset, outperforming \bertlarge~by ~3 F1 points on short answer prediction. 
We also show significant impact in HotpotQA, improving answer prediction F1 by 4 points and supporting fact prediction F1 by 1 point and outperforming the previous best system.
Moreover, we show that our pre-training approach is particularly effective when training data is limited, improving the learning curve by a large amount.

\end{abstract}

\input{intro.tex}
\input{related_work.tex}

\input{bert_background.tex}
\input{sspt.tex}
\input{ds_tasks.tex}
\input{experiments.tex}

\section{Conclusion and Future Work}
\label{sec.conclusion}

Span selection pre-training is effective in improving reading comprehension across four diverse datasets, including both generated and natural questions, and with provided contexts of passages, documents and even passage sets.
This style of pre-training focuses the model on finding semantic connections between two sequences, and supports a style of cloze that can train deep semantic understanding without demanding memorization of specific knowledge in the model.
The span selection task is suitable for pre-training on any domain, since it makes no assumptions about document structure or availability of summary/article pairs. 
This allows pre-training of language understanding models in a very generalizable way.

In future work, we will address end-to-end question answering with pre-training for both the answer selection and retrieval components. 
We hope to progress to a model of general purpose language modeling that uses an indexed long term memory to retrieve world knowledge, rather than holding it in the densely activated transformer encoder layers. 

\FloatBarrier

\bibliography{sspt}
\bibliographystyle{acl_natbib}

\end{document}

%% file: intro.tex
\section{Introduction}

State-of-the-art approaches for NLP tasks are based on language models that are pre-trained on tasks which do not require labeled data \cite{elmo,ulmfit,bert,xlnet,roberta,ernie2}. 
Fine tuning language models to downstream tasks, such as question answering or other natural language understanding tasks, has been shown to be a general and effective strategy. 
BERT is a recently introduced and highly successful model for language understanding.

%
The general BERT adaptation approach is to alter the model used for pre-training while retaining the transformer encoder layers.
The model discards the layers used for the final prediction in the pre-training tasks and adds layers to predict the target task. 
All parameters are then fine tuned on the target task.

BERT is based on the transformer architecture~\cite{NIPS2017_7181}, and trained on the following two unsupervised tasks:
\begin{itemize}
\item Masked Language Model (MLM): predicting masked word pieces from the surrounding context
\item Next Sentence Prediction (NSP): predicting if the two provided sequences follow sequentially in text or not
\end{itemize}

The masked LM or ``cloze'' task~\cite{taylor1953cloze} and next sentence prediction are auxiliary tasks~\cite{ando2005framework} requiring language understanding, and therefore train the model to acquire effective representations of language.
However, the cloze pre-training task often poses instances that require only shallow prediction, or else require memorized knowledge.
For many cloze instances the model simply requires syntactic or lexical understanding to answer.
For example, in the cloze instances in Table \ref{table.clozeExamples} the first two rows require syntactic and lexical understanding respectively. 
Other cloze instances mainly require completing collocations, as in the third example. 
However, some cloze instances require memorized knowledge, as in the last instance, which essentially asks where Hadrian died.

\begin{table}\small
\begin{center}
\setlength\tabcolsep{7pt}
\renewcommand{\arraystretch}{1.8}
\begin{tabular}{rp{0.65\linewidth}}
\toprule Type & Cloze \\ \toprule
Syntactic & In \mask{the} 15th century, the blast furnace spread into what is now Belgium where it was improved. \\
Lexical & Akebia quinata \mask{grows} to 10 m (30 ft) or more in height and has compound leaves with five leaflets. \\
Collocation & Apollo 11 was launched by a Saturn V rocket from Kennedy \mask{Space} Center on Merritt Island, Florida \\
\makecell{Memorized\\Knowledge} & Hadrian died the same year at \mask{Baiae}, and Antoninus had him deified, despite opposition from the Senate. \\
\bottomrule
\end{tabular}
\end{center}
\caption{\label{table.clozeExamples} Cloze instances of different types}
\end{table}

Other language models face the same challenge.
In GPT-2~\cite{gpt2} the entities present in a language generation prompt are expanded with related entities. 
For example, in a prompt about nuclear materials being stolen on a Cincinnati train, GPT-2 references ``Ohio news outlets'', ``U.S. Department of Energy'', and ``Federal Railroad Administration'' in ways consistent with their real world relationships to the entities in the prompt.


As the preceding examples illustrate, in many cloze and conventional language model prediction instances, the correct prediction depends on a specific, narrowly relevant, bit of knowledge.
Further, pre-trained transformer models do indeed encode a substantial number of specific facts in their parameter matrices, enabling them to answer questions directly from the model itself~\cite{gpt2}.
However, because the computational cost of transformers scales at least linearly with the number of parameters, it is expensive to encode all the facts that would enable the correct predictions.
Encoding a large amount of rarely useful information in parameters that are used for every instance is an inefficient use of model capacity if it is not needed for the downstream task.

As the performance gains from GPT to GPT-2 and \bertbase~ to \bertlarge~ show, increasing model capacity continues to provide gains.
Previous work also found seemingly limitless improvements from increasing model capacity~\cite{sparselyGatedExperts}, possible through sparse activation. 
Our hypothesis is that making more efficient use of a fixed number of parameters can provide analogous gains.
In MRC tasks, the model does not need to generate an answer it has encoded in its parameters.
Instead, the task is to use a retrieved passage, or passage set to extract an answer to the question.

To better align the pre-training with the needs of the MRC task, we use \textit{span selection} as an additional auxiliary task.
This task is similar to the cloze task, but is designed to have a fewer simple instances requiring only syntactic or collocation understanding.
For cloze instances that require specific knowledge, rather than training the model to encode this knowledge in its parameterization, we provide a relevant and answer-bearing passage paired with the cloze instance.

We provide an extensive evaluation of the span selection pre-training method across four reading comprehension tasks: the Stanford Question Answering Dataset (SQuAD) in both version 1.1 and 2.0; followed by the Google Natural Questions dataset \cite{Kwiatkowski2019NaturalQA} and a multi-hop Question Answering dataset, HotpotQA~\cite{yang2018hotpotqa}. We report consistent improvements over both \bertbase~and \bertlarge~models in all reading comprehension benchmarks.

The rest of the paper is structured as follows.
In section \ref{sec.relatedWork} We describe earlier work on similar tasks and relate our extended pre-training to the broader research efforts on pre-training transformers. To provide context for our contribution, we review the most relevant parts of BERT in Section \ref{sec.background}.
Next, we describe and formalize our pre-training task and the architectural adjustments to BERT in Section \ref{sec.sspt}.
Finally we provide an extensive empirical evaluation in MRC tasks, describing benchmarks in Section \ref{sec.mrctasks} and evaluating our approach in Section \ref{sec.experiments}. Section \ref{sec.conclusion} concludes the paper highlighting interesting research directiond for future work.

\FloatBarrier

%% file: related_work.tex
\section{Related Work}
\label{sec.relatedWork}


Since the development of BERT there have been many efforts towards adding or modifying the pre-training tasks. \citet{joshi2019spanbert} introduced SpanBERT, a task that predicts the tokens in a span from the boundary token representations. Note that, unlike span selection, there is no relevant passage used to select an answer span. ERNIE 2.0 \cite{ernie2} trained a transformer language model with seven different pre-training tasks, including a variant of masked language model and a generalization of next-sentence prediction.
XLNet \cite{xlnet} introduced the permuted language model task, although it is not clear whether the success of the model is due to the innovative pre-training or larger quantity of pre-training.

In this paper we focus on a pre-training task that has been specifically designed to support QA applications. Previous related work has explored tasks similar to span selection pre-training. These are typically cast as approaches to augment the training data for question answering systems, rather than alleviating the pressure to encode specific facts in the pre-training of a language model.

\citet{NIPS2015_5945} introduces a reading comprehension task constructed automatically from news articles with summaries. In this view the constructed dataset is used both for training and test. Also, entities were replaced with anonymized markers to limit the influence of world knowledge. Unlike our span selection pre-training task, this requires summaries paired with articles and focuses only on entities.
A similar approach was taken in \citet{dhingra2018simple} to augment training data for question answering. Wikipedia articles were divided into introduction and body with sentences from the introduction used to construct queries for the body passage. Phrases and entities are used as possible answer terms.

\citet{onishi2016did} constructed a question answering dataset where answers are always people. Unlike other work, this did not use document structure but instead used a search index to retrieve a related passage for a given question. Because the answers are always people, and there are only a few different people in each passage, the task is multiple choice rather than span selection.
Self training~\cite{selfTrainingQA} has also been used to jointly train to construct questions and generate self-supervised training data.

BERT was trained for one million batches, with 256 token sequences in each. Although this is already a considerable amount of pre-training, recent research has shown continued improvement from additional pre-training data. XLNet \cite{xlnet} used four times as much text, augmenting the Wikipedia and BooksCorpus \cite{bookscorpus} with text from web crawls, the number of instances trained over was also increased by a factor of four. RoBERTa \cite{roberta} enlarged the text corpus by a factor of ten and trained over fifteen times as many instances. This, along with careful tuning of the MLM task resulted in substantial gains. Unfortunately, these very large-scale pre-training approaches require significant hardware resources.  We restrict our experiments to extended pre-training with less than half the steps of BERT (390k batches of 256).


%% file: bert_background.tex
\section{Background}\label{sec.background}
In this section, we give the readers a brief overview of the BERT \cite{bert} pre-training strategy and some details which we modify for our novel span selection auxiliary task.

\subsection{Architecture and setup}
BERT uses a transformer \cite{bert} architecture with $L$ layers and each block uses $A$ self-attention heads with hidden dimension $H$. The input to BERT is a concatenation of two segments  $x_1, \ldots, x_M$ and $y_1, \ldots, y_N$ separated by special delimiter markers like so:  $[CLS],x_1, \ldots, x_M, [SEP], y_1, \ldots, y_N, [SEP]$ such that $M+N < S$ where $S$ is the maximum sequence length allowed during training\footnote{We follow standard notation here as in previous work.}. This is first pre-trained on a large amount of unlabeled data and then fine-tuned on downstream tasks which has labeled data.

\subsection{Objective functions}
BERT used two objective functions during pre-training: masked language modeling and next sentence prediction. We discuss them in brief.\\
\textbf{Masked Language Model (MLM)}: A random sample of the tokens in the input sequence is replaced with a special token called $[MASK]$. MLM computes a cross-entropy loss on predicting these masked tokens. Particularly, BERT selects 15\% of the input tokens uniformly to be replaced. 80\% of these selected tokens are replaced with [MASK] while 10\% are left unchanged, and 10\% are replaced with random token from the vocabulary.\\
\textbf{Next Sentence Prediction (NSP)}: This is a binary classification loss that predicts if two sentences follow each other in the original text. The examples are sampled with equal probability such that positive examples are consecutive sentences while negatives are artificially created by adding sentences from different documents.

%% file: sspt.tex
\section{Span Selection Pre-training}\label{sec.sspt}
In the previous section we briefly discussed the BERT framework along with its objective functions. In this section, we will propose a novel pre-training task for bi-directional language models called span selection.
\subsection{Span Selection}
\label{sec.spanselection}

Span selection is a pre-training task inspired both by the reading comprehension task and the limitations of cloze pre-training.
Figure \ref{spanSelectionExample} illustrates an example of a span selection instance. 
The \textit{query} is a sentence drawn from a corpus with a term replaced with a special token: [BLANK]. 
The term replaced by the blank is the \textit{answer term}.
The \textit{passage} is relevant as determined by a BM25 \cite{bm25} (k1=1.2, b=0.75) search, and answer-bearing (containing the answer term).

\begin{figure}[h]
\begin{center}
\begin{tabular}{p{0.13\linewidth}p{0.77\linewidth}}
\textbf{Query} & ``In a station of the metro'' is an Imagist poem by [BLANK] published in 1913 in the literary magazine Poetry  \\
\textbf{Passage} & \ldots \answer{Ezra Pound}'s  famous Imagist poem, ``In a station of the metro'', was inspired by this station \ldots \\
\textbf{Answer Term} & Ezra Pound \\
\end{tabular}
\end{center}
\caption{\label{spanSelectionExample} Example Span Selection Instance}
\end{figure}

Unlike BERT's cloze task, where the answer must be drawn from the model itself, the answer is found in a passage using language understanding.

\begin{figure}[h]
   \centering
   \includegraphics[width=\linewidth]{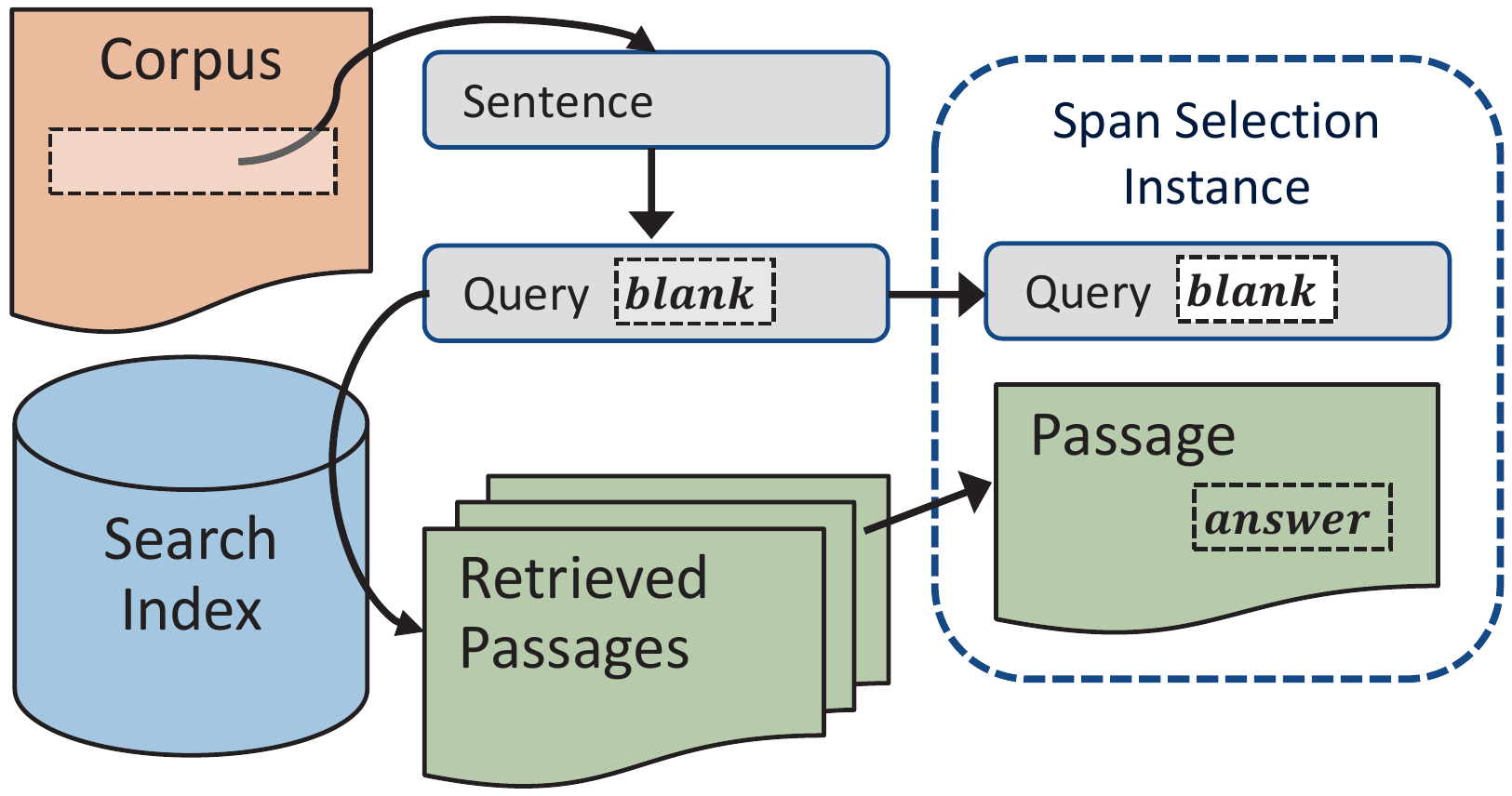}
   \caption{Span Selection Training Generation}
   \label{fig.ssptGeneration}
\end{figure}

Figure \ref{fig.ssptGeneration} outlines the process of generating span selection pre-training data.
The input is an unlabeled corpus, which is then split into passages and indexed.
We used passages from Wikipedia\footnote{December 2018 snapshot} 300 to 2000 characters long, split on paragraph boundaries, and Lucene\footnote{\url{http://lucene.apache.org/}} 7.4.0 as the search engine.
In addition to the text of the passage, we store the document ID, so that we may filter passages that occur in the same document as the query.

To gather queries, we iterate over the sentences in the corpus between 50 and 250 characters long. For each sentence, we choose an answer term to replace with a blank.
We used a set of simple heuristic criteria to identify answer terms that are likely to result in queries that require deep understanding to answer: the term should be between 4 and 30 characters and either a single token from an open class part-of-speech (20\%) or a noun phrase or entity (80\%), as detected by a part-of-speech pattern and ClearNLP NER.

To identify the passages, we use the generated query, with the answer term removed, as a bag-of-words query to search into the passage index.
The top ten results were searched for an answer-bearing passage; if none were found the query was either discarded or sampled to maintain a 30\% composition of \textit{impossible} span selection instances.
The impossible instances are those that do not have the answer-term in the provided passage.
We further required a minimum BM25 score of 25 (tuned manually to reflect high relevance).
If the answer term was part of a longer sequence of tokens shared by the query and passage, we extended the answer term to be the longest such sequence.
This avoids cases where the answer term can be found through trivial surface-level matching.


\begin{table}\small
\begin{center}
\begin{tabular}{rp{0.7\linewidth}}
\toprule Type & Span Selection Instance \\ \toprule
\makecell[tr]{Phrase\\Multiple\\Choice} & \textbf{Q:} The year 1994 was proclaimed [BLANK] of the Family by the United Nations General Assembly.\\
\vspace{0.2cm}
& \textbf{P:} \answer{The International Year} for the Culture of Peace was designated by the United Nations as the year 2000, with the aim of celebrating and encouraging a culture of peace. \ldots \\

\makecell[tr]{Suggestive\\Inference} & \textbf{Q:} On the island of Kaja in [BLANK], a male orangutan was observed using a pole apparently trying to spear or bludgeon fish.\\
\vspace{0.2cm}
& \textbf{P:} \ldots Although similar swamps can be found in \answer{Borneo}, wild Bornean orangutans have not been seen using these types of tools. \\

\makecell[tr]{Justified\\Inference} & \textbf{Q:} The company's headquarters are located in the city of Redlands, California, 50 miles east of [BLANK].\\
\vspace{0.2cm}
& \textbf{P:} Redlands (Serrano: Tukut) is a city in San Bernardino County, California, United States. It is a part of the Greater \answer{Los Angeles} area. \ldots \\
\bottomrule
\end{tabular}
\end{center}
\caption{\label{table.ssptTypes} Span Selection instances of different types}
\end{table}

Table \ref{table.ssptTypes} shows examples of span selection instances of different types.
Rather than discreet types, these are best understood as a continuum.
Comparing to the cloze types in Table \ref{table.clozeExamples}, we see an analogy between the lexical cloze type and phrase multiple choice.
These two types involve understanding what words (or phrases) are reasonable in the context from the set of wordpieces (or possible spans).
The memorized knowledge cloze type contrasts with the suggestive or justified inference span selection types.
Because a suggestive or justifying passage is present, the model is trained only to understand language, rather than memorize facts. 
Simple syntactic instances are largely eliminated because closed class words are not possible answer terms.
Also, since answer terms are expanded to the longest shared subsequence between query and passage, collocation instances are not a concern.

\subsection{Extended Pre-training}
Rather than training a transformer architecture from scratch, we initialize from the pre-trained BERT models \cite{bert} and extend the pre-training with the span selection auxiliary task. 
We refer to the resulting models as \bertbase+SSPT (Span Selection Pre-Training) and \bertlarge+SSPT.
We used batch sizes of 256, and a learn rate of 5e-5. All models were trained over 100 million span selection instances. We found continued improvement from 50 million to 100 million and have not yet tried larger pre-training runs.
Unlike the efforts of XLNet or RoBERTa which increased training by a factor of ten relative to BERT, the additional data in SSPT represents less than a 40\% increase in the pre-training of the transformer.
This pre-training is also done over Wikipedia, adding no new text to the pre-training.

\begin{figure}[b]
   \centering
   \includegraphics[width=\linewidth]{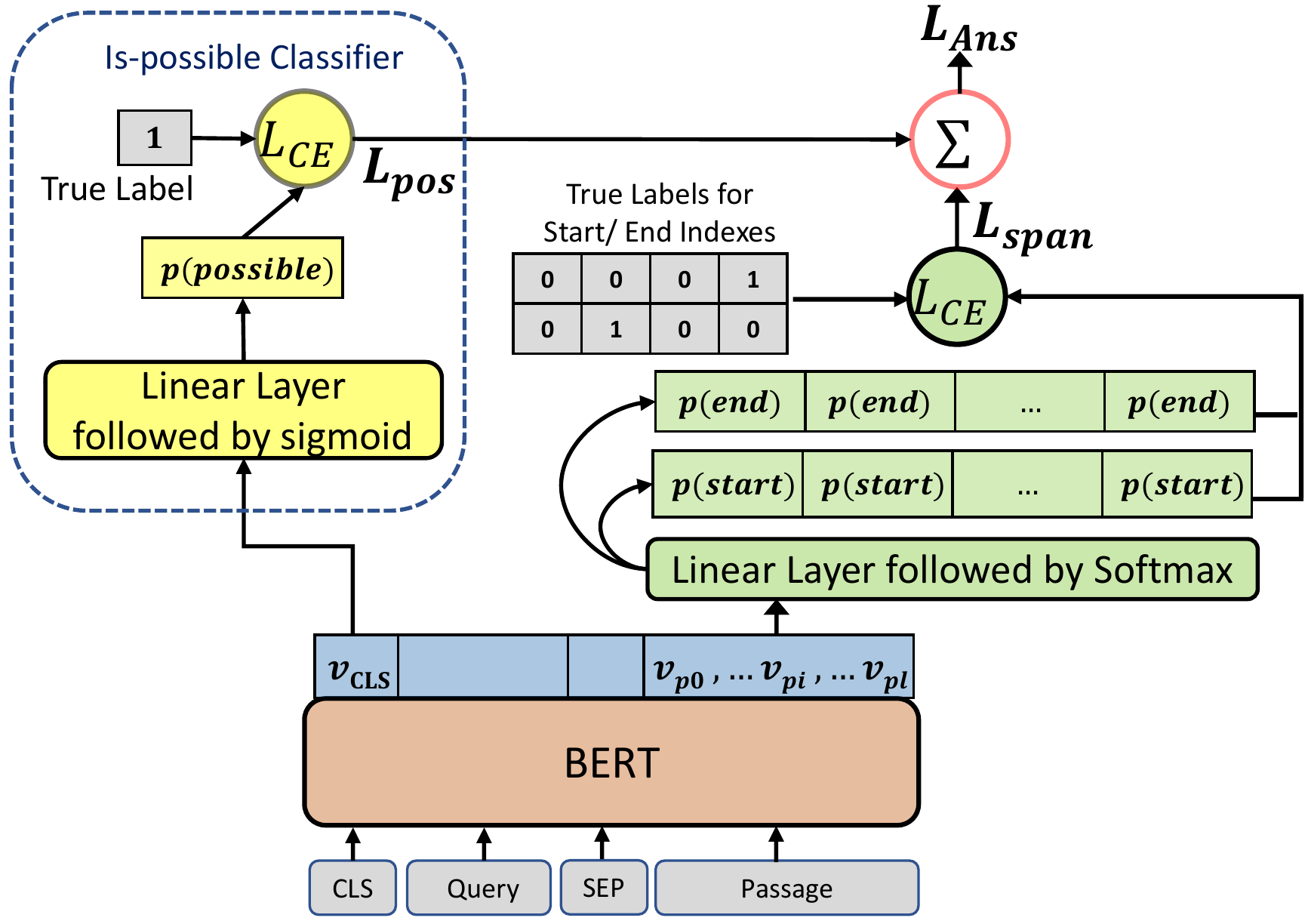}
   \caption{BERT for QA with is-possible prediction}
   \label{fig.bertForQA}
\end{figure}

Figure \ref{fig.bertForQA} illustrates the adaptation of BERT to SSPT.
The query and passage are concatenated in the standard two sequence representation, with a preceding [CLS] token and a separating [SEP] token, producing a sequence of tokens $T$.
BERT produces output vectors for these tokens to obtain a sequence $\{\boldsymbol{v}_i\}_{i=1}^{|T|}$ of $d$ dimensional vectors.

In span selection extended pre-training, we alter the vocabulary of the tokenizer, introducing the new special token: `[BLANK]'. 
We use the BertForQuestionAnswering\footnote{\url{https://github.com/huggingface/pytorch-transformers}} model, which uses a pointer network to find the answer location.
The pointer network applies a simple fully connected network to predict the probability of start and end span pointers at each token position, using the output of the final transformer layer at that position.
The loss in training is the cross entropy of these predictions with the true positions of the start and end.

Formally, The start of the answer span is predicted as $p(i=\langle start \rangle) = softmax(\boldsymbol{w}_{\langle start \rangle}^{\top}\boldsymbol{v} + b_{\langle start \rangle})_i$, 
where $\boldsymbol{w}_{\langle start \rangle}\in {\mathbb{R}}^{d}$, $b_{\langle start \rangle} \in \mathbb{R}$ are trainable parameters. 
Then end of the span is predicted the same way: $p(i=\langle end \rangle) = softmax(\boldsymbol{w}_{\langle end \rangle}^{\top}\boldsymbol{v} + b_{\langle end \rangle})_i$.

\begin{table*}[th]
\begin{center}
\begin{tabular*}{0.9\textwidth}{@{\extracolsep{\fill}}cccccccc}
\toprule 
\thead{\textbf{Dataset}} &  \thead{\textbf{Context}} & \thead{\textbf{Answer Types}} & \thead{\textbf{Question}\\ \textbf{Creation}} & \thead{\textbf{Training}\\ \textbf{Size}} & \thead{\textbf{Dev}\\ \textbf{Size}} & \thead{\textbf{Test}\\ \textbf{Size}} & \thead{\textbf{Gap to Human} \\ \textbf{Performance}\textdagger} \\ 
\toprule
\vspace{0.1cm}
SQuAD 1.1         & passage   & span & generated & 88k & 11k & 10k & $<0$\% \\
\vspace{0.1cm}
SQuAD 2.0         & passage  & \makecell{span,\\impossible} & generated & 130k & 12k & 9k & $<0$\% \\
\vspace{0.1cm}
\makecell{Natural\\Questions} & document   & \makecell{span,yes,no,\\impossible} & natural & 307k & 8k & 8k & 15\%\\
\vspace{0.1cm}
HotpotQA          & passage set & \makecell{span,yes,no} & generated & 91k & 7k & 7k & 8\% \\
\bottomrule
\end{tabular*}
\end{center}  
\caption{\label{tbl.qadatasets} Comparison of QA Datasets. \hspace{4cm} \textdagger As of Dec. 2019 }
\end{table*}

Span selection pre-training may optionally include a classifier for answerability. 
If the answerability classifier is included in the pre-training then the presence of the answer span in the passage is predicted with probability given by: $p(possible)=sigmoid(\boldsymbol{w}_{CLS}^{\top} \boldsymbol{v}_{CLS} + b_{CLS})$. 
If it is not included, for impossible instances the target prediction is for both start and end to be position zero, the [CLS] token. 
We train models for QA without the answerability classifier for 100 million instances.  This took approximately seven days on 16 P100 GPUs.

Training data and code to extend pre-training is available as open source\footnote{\url{https://github.com/IBM/span-selection-pretraining}}.



%% file: ds_tasks.tex
\section{MRC Tasks}\label{sec.mrctasks}

We follow previous work and evaluate our SSPT architecture on several downstream tasks. Our primary motivation is to improve question answering by improving the pre-trained language model. Our QA benchmarks are the following:
\begin{enumerate}
    \item Stanford Question Answering Dataset (SQuAD) v1.1 \cite{squad} and v2.0 \cite{squad2}
    \item Natural Questions (NQ) \cite{Kwiatkowski2019NaturalQA}
    \item HotpotQA \cite{yang2018hotpotqa}
\end{enumerate}
The three datasets provide different characteristics of question answering and machine reading comprehension tasks as well as an opportunity to compare results with active leaderboards.
Table \ref{tbl.qadatasets} provides a summary comparison.
We briefly discuss them here:

\subsection{SQuAD}
SQuAD provides a paragraph of context and asks several questions about it. The task is extractive QA where the system must find the span of the correct answer from the context. We evaluate on two versions of SQuAD: v1.1 and v2.0. In v1.1 the context always contains an answer. However, in v2.0 the task contains additional questions to which the given context does not have the correct answer.

Just as in Figure \ref{fig.bertForQA}, the question and passage are concatenated with the separators ([CLS] and [SEP]) to form the input to the pre-trained BERT. The final token representations are then used to predict the probability for each token that it is the start or end of the answer span. The span with the highest predicted probability is then the predicted answer.

\subsection{Natural Questions} NQ is a dataset of over 300,000 queries sampled from live users on the Google search engine for which a Wikipedia article is contained in the top ranking search results.
Crowd sourced annotators are then tasked with highlighting a short answer span to each question\footnote{Around 1\% of the questions are answered as a simple Yes or No rather than a span of short answer text.
Due to their small proportion, the models in this paper do not produce Yes/No answers}, if available, from the Wikipedia article as well as a long answer span (which is generally the most immediate HTML paragraph, list, or table span containing the short answer span), if available.

Similar to SQuAD 2.0 the NQ dataset forces models to make an attempt at ``knowing what they don't know'' in order to detect and avoid providing answers to unanswerable questions.  In addition, the fact that the questions were encountered naturally from actual users removes some of the observational bias that appears in the artificially created SQuAD questions. Both these aspects along with the recency of the task's publication means that this is still a challenging task with lots of headroom between human performance and the best performing automated system.\\

Experiments on the NQ dataset use the strategies and model described by \citet{alberti2019bert} to fine tune a \bertlarge{} model with a final layer for answerability prediction as well as sequence start/end prediction.
Similar to their best performing systems, the model is first trained using the SQuAD v1.1 data set and then subsequently trained on the NQ task\footnote{Skipping the SQuAD v1.1 fine-tuning step for the NQ task leads to the same conclusions with respect to SSPT pre-training, but decreases the overall performance for both \bertlarge{} and \bertlarge{}+SSPT}.
The hyperparameters follow \citet{alberti2019bert} with the exception of learning rate and batch size which are chosen according to the approach outlined by \citet{smith2018disciplined} using a 20\% sub-sample of the data for each experimental setting.

\subsection{HotpotQA} 
Recently, \citet{yang2018hotpotqa} released a new dataset, called HotpotQA, for the task of reading comprehension style extractive QA.
Each training instance in the {\em distractor} setting of this dataset comprises a question, a set of ten passages, an answer, and a binary label for each sentence in the passage-set stating whether that sentence serves as a supporting fact (or not) to arrive at the correct answer.
The task is to predict both the correct answer as well as the supporting facts for any given test instance.
The signature characteristic of this dataset lies in the fact that each question requires a minimum of two supporting facts from two different passages in order to derive its correct answer.
Thus, this dataset tests the cross-passage, multi-hop reasoning capability of a reading comprehension based question answering system.

Our system for HotpotQA uses a three-phase approach. First, representations of the individual passages are built with a pre-trained transformer encoder. Second, interactions between these passages are attended to using a relatively shallow {\em global} transformer encoder. The supporting facts are predicted from the sentence representations produced by this global layer.
Finally, the predicted supporting facts are then merged into a {\em pseudo-passage} that is used by a slightly altered version of the model for SQuAD. The one addition is that this model also predicts an answer-type ($\{yes, no, span\}$) from the [CLS] token vector.

%% file: experiments.tex
\section{Experiments}
\label{sec.experiments}

\begin{table}[t!]
\begin{center}
\resizebox{\linewidth}{!}{%
\begin{tabular}{rccccc}
\toprule \textbf{Method} & \multicolumn{2}{c}{\textbf{SQUAD 1.1}}  && \multicolumn{2}{c}{\textbf{SQUAD 2.0}} \\
\cline{2-3} \cline{5-6}           &  F1    & Exact  && F1     & Exact  \\ \toprule
\bertbase         & 88.52 &	81.22 && 76.45 & 73.29 \\
+SSPT    & \best{91.71} &	\best{85.10} && 82.31 & 79.19 \\
+SSPT-PN & 91.60 &  84.94 && \best{82.34} & \best{79.32} \\
\hline
\bertlarge        & 90.97 &	84.20 && 81.50 & 78.41  \\
+SSPT   & \best{92.75} &	\best{86.86} && \best{85.03} & \best{82.07}  \\
\bottomrule
\end{tabular}}
\end{center}
\caption{\label{tbl.squad} Dev Set Results on SQuAD}
\end{table}

\avi{please add the Aparikh and Alberti paper numbers to compare with us. So we'll have 2 more rows to compare against. if 52.7 is theirs, then we will have 1 more row to compare against. }
\begin{table}[t!]
\begin{center}
\begin{tabular}{rcc}
\toprule \textbf{Method} & Short Ans F1 & Long Ans F1 \\ \toprule
\bertbase          & 47.27 & 61.02 \\
+SSPT    & \best{50.40} & \best{63.35} \\ \hline
\bertlarge          & 52.7 & 64.7 \\
+SSPT    & \best{54.2} & \best{65.85} \\
\bottomrule
\end{tabular}
\end{center}
\caption{\label{tbl.naturalquestion} Dev Set Results on Natural Questions}
\end{table}

\begin{table}[t!]
\begin{center}
\resizebox{\linewidth}{!}{%
\begin{tabular}{rccccc}
\toprule \textbf{Method} & \multicolumn{2}{c}{Facts} && \multicolumn{2}{c}{Answer} \\ \cline{2-3}\cline{5-6}
& F1 & Exact && F1 & Exact \\ \toprule
\bertbase        & 84.00 & 53.15 && 73.86 & 59.97 \\
+SSPT   & \best{85.13} & \best{56.58} && \best{77.25} & \best{63.31} \\ \hline
\bertlarge       & 85.27  & 55.99 && 75.48 & 61.62 \\
+SSPT  & \best{86.17} & \best{57.57} && \best{79.39} & \best{65.87} \\
\bottomrule
\end{tabular}}
\end{center}
\caption{\label{tbl.hotpotqa} Dev Set Results on HotpotQA}
\end{table}

\begin{figure}[ht]
   \centering
   \includegraphics[width=\linewidth]{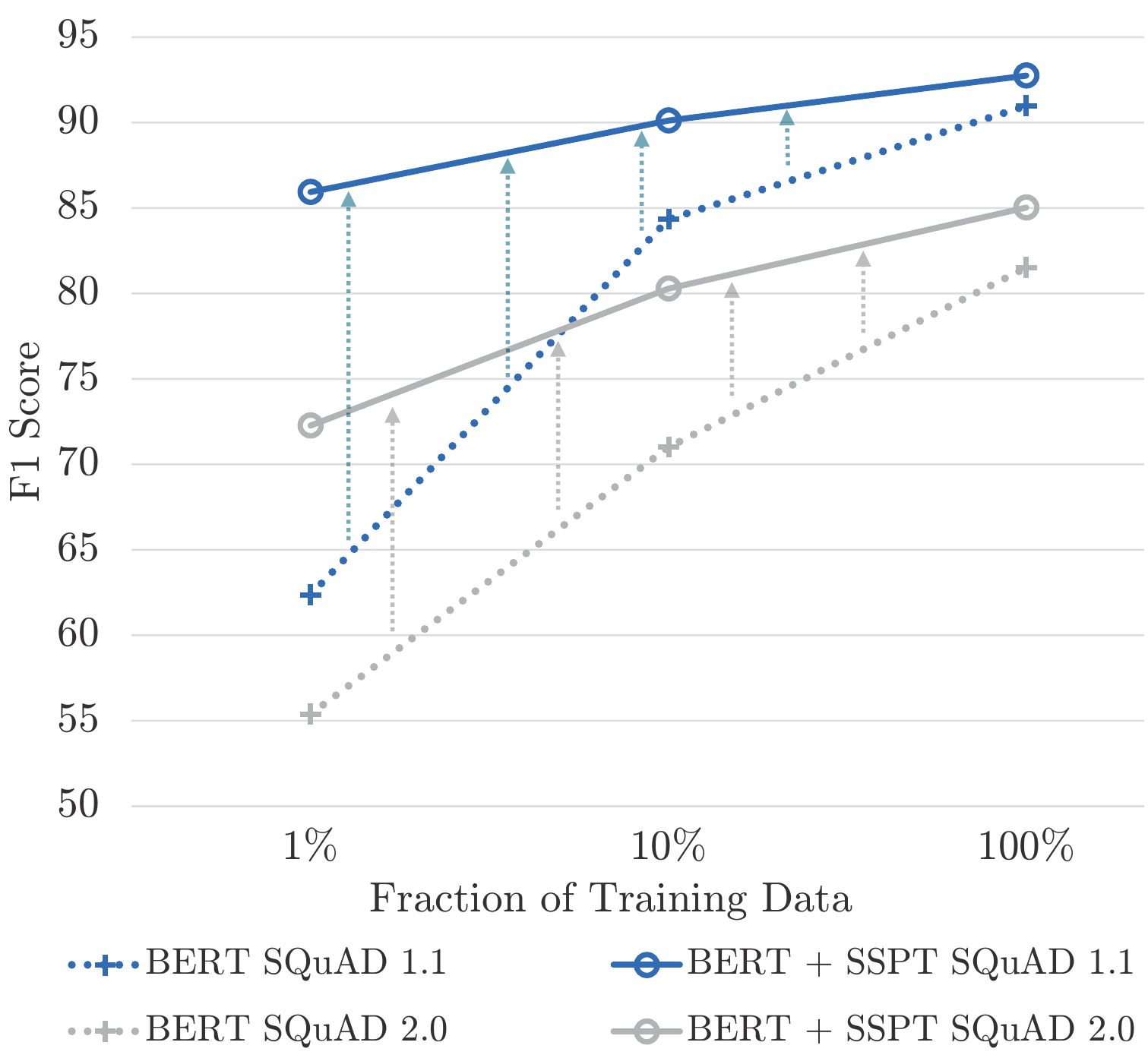}
   \caption{Learning curve improvement for \bertlarge~with SSPT}
   \label{fig.learningCurve}
\end{figure}

Tables \ref{tbl.squad}, \ref{tbl.naturalquestion}, and \ref{tbl.hotpotqa} show our results on the development set with extended span selection pre-training for BERT relative to the pre-trained BERT.
We use the same hyperparameters on these tasks as the original BERT.
The best results for each dataset are in bold when significant relative to the BERT baseline.
The four question answering datasets are improved substantially with span selection pre-training.

\subsection{SQuAD}

Relative to \bertbase{} we find a 3 point improvement in F1 for SQuAD 1.1 and a nearly 6 point improvement for SQuAD 2.0. In terms of error rate reduction the improvement is similar, 28\% and 25\% respectively. The error rate reduction for \bertlarge{} is 20\% and 19\% for SQuAD 1.1 and 2.0 respectively.

In reading comprehension tasks, the pointer network for answer selection is pre-trained through the span selection task.
\mrglass{we explore this effect on the SQuAD dataset (but we don't repeat for others)}
We measure how much of the improvement is due to this final layer pre-training versus the extended pre-training for the transformer encoder layers by discarding the pre-trained pointer network and randomly initializing.
This configuration is indicated as \bertbase+SSPT-PN.
Surprisingly, the pre-training of the pointer network is not a significant factor in the improved performance on reading comprehension, indicating the improvement is instead coming through a better language understanding in the transformer.

Figure \ref{fig.learningCurve} shows the improvement from SSPT on SQuAD 1.1 and 2.0 as the amount of training data increases. 
While there is significant improvement at 100\% training, the improvement is even more pronounced with less training data.
We hypothesize that this is due to the close connection of span selection pre-training with reading comprehension.
This effect is strongest for SQuAD 1.1, which like span selection pre-training always contains a correct answer span in the passage.


\subsection{Natural Questions}
The work of \citet{albert-synth-data}, which gets the \bertlarge{}  performance listed in Table \ref{tbl.naturalquestion}, is the highest ranking single model submission that does not use data augmentation with a published paper. Our implementation of \bertlarge{}+SSPT, therefore, provides a 1.5\% improvement over the best BERT-for-QA model performance that we are aware of on the NQ data set. In future work, we intend to explore data augmentation on top of \bertlarge{}+SSPT for further improvements.

\subsection{HotpotQA}
In HotpotQA, unlike the other QA datasets, multiple passages are provided.
We use the BERT transformer in two places, for supporting fact prediction to build the representations of each passage, and in answer prediction as in the other QA tasks.
We find the most substantial gains of almost 4 F1 points for answer selection, the QA task most similar to span selection pre-training. 
Interestingly, we also find improvement of almost one point F1 in supporting fact prediction, demonstrating that the learned representations can generalize well to multiple QA sub-tasks.

HotpotQA also comes with its own leaderboard (https://hotpotqa.github.io/). A good number of submissions on this leaderboard are based on \bertbase{}  or \bertlarge. 
We made an initial submission to this leaderboard, called TAP, which occupied Rank-5 at the time of submission and the underlying architecture employed \bertbase{}. 
Next, we replaced \bertbase{} with \bertlarge{}+SSPT, calling that model TAP-2. This change resulted in a $7.22\%$ absolute gain in the Joint $F1$ score. 
An ensemble version of TAP-2 further offered a gain of $1.53\%$.
The SSPT augmented TAP-2 (ensemble) and TAP-2 (single model) achieved Rank-1 and Rank-2 on the leaderboard at the time of submission. 

\subsection{Exploration of SSPT Instance Types}
In section \ref{sec.spanselection} we enumerated three types of span selection instances. The first type, Phrase Multiple Choice, is the least interesting since the semantic correspondence between the query and the passage is not used. Instead, the instance is treated as a cloze with options provided as spans in the passage.  Note that in this type of instance the relevance of the passage to the query is not important.

To explore how frequent this case might be we select 100 thousand new SSPT instances with a relevant passage and for each select an alternative, random, answer-bearing, passage. The unrelated passage is from a document  different both from the query's document and from the relevant passage's document. We then apply the SSPT trained model to the instances both with the related and unrelated passage and evaluate its performance in terms of token-level F1 and exact span match.

Table \ref{tbl.randomPassage} show the performance of our SSPT trained models on the SSPT queries with related vs. unrelated passages.  The large accuracy gains when using relevant passages imply that for many passages ``Phrase Multiple Choice'' is not the method used by the model. Instead, the semantic connection of the passage to the query is used to select the appropriate span.


\begin{table}[t!]
\begin{center}
\resizebox{\linewidth}{!}{%
\begin{tabular}{rrcc}
\toprule \textbf{Model} & \textbf{Passage} & F1 & Exact \\ \toprule
\bertbase+SSPT  & Related   & \best{62.88} & \best{49.27} \\
\bertbase+SSPT  & Unrelated & 46.51 & 34.32 \\ \hline
\bertlarge+SSPT & Related   & \best{65.39} & \best{51.82} \\
\bertlarge+SSPT & Unrelated & 50.98 & 38.97 \\
\bottomrule
\end{tabular}
}
\end{center}
\caption{\label{tbl.randomPassage} Comparison of performance of SSPT for related vs. unrelated passages}
\end{table}

\subsection{Comparison to Previous Work}
We also compare our span selection pre-training data with the data distributed by \citet{dhingra2018simple}. This data consists of approximately 2 million instances constructed using the abstract and body structure of Wikipedia. In contrast, our approach to pre-training can generate data in unlimited quantity from any text source without assuming a particular document structure. 
When only one million training steps are used, both sources of pre-training are equally effective. But when moving to ten million steps of training, our data produces models that give over one percent better F1 on both SQuAD 1.1 and 2.0.
This suggests the greater quantity of data possible through SSPT is a powerful advantage.